# MANorm: A Normalization Dictionary for Moroccan Arabic Dialect Written in Latin Script

**Randa Zarnoufi**
FST, Department of Computer Science
University Mohammed V in Rabat
randa_zarnoufi@um5.ac.ma

**Walid Bachri**
ENSIAS
University Mohammed V in Rabat
bachriwalid@gmail.com

**Hamid Jaafar**
ENS, Department of Languages and Educational Sciences
University Hassan II
Jaafarhamid1973@gmail.com

**Mounia Abik**
ENSIAS
University Mohammed V in Rabat
mounia.abik@um5.ac.ma

## Abstract

Social media user generated text is actually the main resource for many NLP tasks. This text however, does not follow the standard rules of writing. Moreover, the use of dialect such as Moroccan Arabic in written communications increases further NLP tasks complexity. A dialect is a verbal language that does not have a standard orthography, which leads users to improvise spelling while writing. Thus, for the same word we can find multiple forms of transliterations. Subsequently, it is mandatory to normalize these different transliterations to one canonical word form. To reach this goal, we have exploited the powerfulness of word embedding models generated with a corpus of YouTube comments. Besides, using a Moroccan Arabic dialect dictionary that provides the canonical forms, we have built a normalization dictionary that we refer to as *MANorm*[1]. We have conducted several experiments to demonstrate the efficiency of MANorm, which have shown its usefulness in dialect normalization.

## 1 Introduction

The large part of the world's population is daily connected and very active in Social Media (SM). This community produces a huge amount of data. This latter, especially textual one is actually very useful for the development of many NLP or text-based applications in general (Farzindar and Inkpen, 2018). However, these texts generated by SM users are of a noisy nature or in other words do not follow the rules of standard communications. Another phenomenon, which adds more complexity to this type of content, is the use of dialects, which are non-standard languages used mainly in verbal communication.

Since the advent of Short Messaging Service (SMS), Moroccan Arabic (MA) dialect has been introduced into users written communications[2] and today, in social media, this phenomenon is becoming widespread (Caubet, 2017). MA dialect is the mother tongue of most Moroccan people, it has been used in SM to freely and spontaneously express emotions and thoughts with other peers (Hall, 2015). This language does not have a standard spelling since it is not used as formal language. Therefore, each social media user writes according to his own. The writing variability is due to the diversity of individual's pronunciation related to their different regional and cultural backgrounds (Boukous, 1995). Thus, for the same word, we find different spellings. For instance, the word 'chkoun' (who) has other five different transliterations ('chkoune', 'chkon,' chkone', 'chkou', 'chkoon'). This problem constitutes a major handicap for many NLP tasks (Han and Baldwin, 2011). To overcome this problem, normalization can be used as a preprocessing in front of the main NLP task. This preprocessing has proved his efficiency in sentiment analysis (Htait et al., 2018), dependency parsing (Van Der Goot et al., 2020) and also in POS tagging (Bhat et al., 2018). In a previous work (Zarnoufi et al., 2020), we have introduced Machine Normalization system for social media text standardization, that can be used as a preprocessing. The current work will be part of this system which will allow us to improve its performance.

---

[1] MANorm is available at: //github.com/MAProcessing/MANorm
[2] Until 1998, writing in MA was not yet recognized, with the exception of a few essays from literacy classes and some songs such as "Melhoun" and some linguistic resources (Jaafar, 2012).

For standard language, in general the task of normalization aims at mapping each out of vocabulary word to one correct form or standard form among a set of standard words candidates (n→1). On the contrary, for dialect, which is a non-standard language. Since there is no standard form for dialect words, this task starts from considering a transliteration of word phonemes (the words are written as spoken) as the canonical form. Then, we try to capture all its possible transliteration forms (1→n). In this work, we follow this approach for MA dialect normalization. First, we build a MA words dictionary that we consider as the lexicon of canonical word form. We then exploit distributed word representations models trained on a YouTube comments corpus to extract the most similar (semantically) words of each dictionary entry, and lexical similarity measures to select all the nearest word forms. The result is a normalization dictionary mapping between each MA word transliteration and its canonical form. We refer to the constructed normalization dictionary for Moroccan Arabic as MANorm.

In the next section, we discuss related works particularly closely related ones on standard languages. Then, we present the detailed solution with the used resources followed by the evaluation and the discussion of the resulted dictionary and we conclude with future directions and further challenges.

## 2 Related Works

Text normalization can be seen as the successor of spelling correction, with the difference that in the first case, the noisy writing style is often intentional (Han and Baldwin, 2011), for example when using abbreviations (e.g. use of 'u' instead of 'you'). Whereas in the second case, misspellings are unintentional and are mainly related to cognitive processes (J.Steffler, 2001). The first approaches for text normalization were based on rules and the noisy channel model (Shannon, 1948) mainly related to spelling correction techniques. They were used jointly for automatic normalization. Among these rules, we find lexical rules using edit distance[3] (Sidarenka et al., 2013) that can detect misspellings and regular expression patterns used for removing or replacing unnecessary character repetitions or URLs, hashtags and logograms[4]. They can also be used for the detection of SM special words (Cotelo et al., 2015). In addition, phonetic rules using Soundex algorithm variants, can serve to normalize noisy word related to pronunciation differences (Eryigit and Torunoglu-Selamet, 2017). The noisy channel model normalizes words by selecting the most probable formal ones using probabilities ranking from language model. It was used with supervised and unsupervised training (Cook and Stevenson, 2009). In general, these approaches capture the differences of the word's surface forms by detecting the similarity in the lexical level between informal and formal word forms. However, the semantic level remains inaccessible because these techniques are not able to capture the words' context. The problem here is that an informal word can be assigned to a formal one, only based on its lexical form without considering its meaning, which can constitute a source of ambiguity.

To mitigate this drawback, supervised learning, machine translation and other techniques were used. For supervised learning, features such as character N-grams, word embedding, POS tag, edit-distances, lookup lists and others are used with labeled data as in MoNoise (Van Der Goot and Van Noord, 2017), which is the current state-of-the-art model for most languages. In addition, different architectures of neural networks were adopted such as LSTM (long short-term memory) model to predict the word canonical form, using the word itself and its surrounding words as in (Min and Mott, 2015). In a very recent work, Muller et al. (2019) have tried to use contextualized embedding with BERT (Bidirectional Encoder Representations from Transformers) to learn lexical normalization for English. This task has also been approached as statistical machine translation SMT-like task (Kaufmann and Kalita, 2010) as a means of context-sensitive technique, where the goal has been to translate noisy text into standard one using parallel corpora. CSMT or character level SMT has also been used for normalization and performed better results than word level SMT (Scherrer and Ljubešić, 2016). Neural machine translation (NMT) was also used, in Lusetti et al. (2018) a neural encoder-decoder with word and character level language model surpassed CSMT performance. Nevertheless, these techniques require a large scale of labeled data, which is in itself a complex and costly task.

---

[3] Edit distance is the number of applied operations to transform one string into another. It allows measuring the lexical similarity between strings. Levenshtein distance (Levenshtein, 1966) is the most used measure that includes insertion, deletion and substitution operations.

[4] Using a single letter or number to represent a word or word part.

To address these problems, Sridhar (2015) was the first to introduce the contextualized normalization with a fully unsupervised manner. He has employed the distributed representation of words or word embedding to capture contextual similarity that can match the noisy word to its canonical form if they share the same vector representation. In other words, their vectors are the closest to each other among all the vocabulary. He has used finite state machines (FSM) to represent the resulting lexicon, and the normalization process is carried out by transducing the noisy words from the FSM. The main advantages of this technique are, first, the needless of labeled corpus, he has used Twitter and customer care notes as training data therefore it is scalable and adaptive to any language. Second, the presence of the contextual dimension, which has been a key factor of its high performance in this task that surpassed Microsoft Word and Aspell accuracies.

These positive qualities have inspired other works on normalization. Bertaglia and Nunes (2016) have performed Portuguese normalization using word embedding model trained on products reviews and tweets. They have built a dictionary mapping between noisy and canonical words to represent the lexicon. They have conduct experiments on both internet slang and orthographic error correction. The obtained results outperformed existing tools. Htait and Bellot (2018) have employed the same approach to build normalization dictionaries for English, French and Arabic using Twitter corpora to overcome the lack of normalization resources available for these languages. They have also reached high performance in the three languages.

All these mentioned works have been done for standard languages normalization. For dialects that suffer from resources scarceness, the related works are very limited. Among them we find Conventional Orthography for Dialectal Arabic or CODA, first proposed in (Habash et al., 2012) for the Egyptian dialect. It was later improved to CODA* in (Habash et al., 2018) and extended to include new dialects. Al-badrashiny et al. (2014), have introduced a system that generates a list of all possible transliterations for each word in an input sentence using a finite-state transducer trained on character-level alignment from Egyptian dialect written in Arabizi (Latin script) to Arabic script. They have learned the transducer on parallel corpus of Egyptian Arabizi-Arabic words. Partanen et al. (2019) have used character level NMT to translate dialectal Finnish to standard one. They used LSTM and transformer models that has been trained on a hand-annotated corpus of transcriptions of different speech records starting from 1950. Word embedding has also been used in dialect processing for the construction of a comparable corpus and a lexicon of Algerian dialect (Abidi and Smaili, 2018) by alignment of a corpus extracted from YouTube. The built lexicon associates the different transliterations forms of dialect words written in Arabic and Latin scripts.

As a dialect, Moroccan Arabic is an under-resourced language. There is no available resource or tool for its normalization. The only work done for this purpose was (Tachicart and Bouzoubaa, 2019) where the authors have used a corpus from Facebook and YouTube to analyze spelling inconsistency of MA dialect used in SM text written in Arabic and Latin scripts. They have compared their corpus with a reference dictionary that has been previously built by the authors and they have found that 35% of this text is noisy. They have concluded that a spell-correction tool is essential to clean up and convert dialectal words into a single standard form of writing.

In line with previous works, namely Sridhar and Htait, we will employ distributed representation for our MA normalization for multiple reasons. First, we only need a corpus of unlabeled data and a dictionary of dialect words to serve as a lexicon for normalized word forms. Second, word embedding is able to identify semantically similar words because it constructs word vectors based on the assumption that semantically similar words are surrounded by the same context. Therefore, the semantic aspect of each reference word and its associated words is guaranteed. In other words, the reference word and these extracted synonyms have the same meaning in the context in use. The remaining task is then to measure the lexical similarity between these words to identify the different lexical forms of the same canonical word. Finally, a normalized form is provided to these words.

## 3 Moroccan Arabic Dialect words forms

The MA dialect is mostly derived from Arabic[5] about 86% and a mixture of other languages namely, French 11.72%, Tamazight 0.39% and Spanish 0.06% according to (Tachicart et al., 2016). As previously mentioned, we are interested in MA dialect normalization, specially the dialect written in Latin script (also called Arabizi). The first time this script appeared goes back to the beginning of SMS in early 2000's where mobile phones did not yet have an Arabic keypad and some phones cannot display messages written in Arabic script, whereas Latin script was accessible in all phones. However, until today, there are still people who are keeping this type of writing even if Arabic keypads are widely available. The MA dialect written in Latin script is the transliteration of phonemes mainly of Arabic origin, thus it is more speech like than writing like. This script uses Latin consonants that mimic Arabic ones and vowels as the equivalent of diacritics.

To identify the different phenomena that dialect normalization needs to address. We selected some comments from our collected corpus composed of YouTube comments (see Sec. 4). Then we analyzed the words forms to determine the sources of lexical variation. We identified five categories as listed below:

- *Vowels variants for the same phoneme*: this is mainly due to pronunciation differences between regions. For example, 'a' and 'e' may be used interchangeably as in 'bayan' and 'bayen' (clear). We have also observed that vowels may be omitted in some cases like in 'm3alqa' and 'm3lqa' (spoon).

- *Letter substitution by number*: some number are used instead of letter to represent Arabic grapheme, if their graphical form is close to a letter in Arabic script. For example, the use of '9' rather than 'ق' [q] and '7' instead of 'ح' [ḥ]. The detailed cases are presented in Table 1.

- *Gemination*[6]: is frequent in Arabic, and it is represented by double consonants that are mentioned by some users and overlooked by others. For example, 'm3allam' (skillful) may be written 'm3alam'.

- *Words combination*: the words in some specific phrases are combined to form one word. For example, "hamdo li allah" (thanks god) may be written for example as 'hamdoulillah' or 'hamdollah' or 'hamdouallah'.

- *Word agglutination*: MA is mostly derived from Arabic, which is highly inflectional or agglutinative language where affixes are combined with the main word. For example, the expression 'wlidatou' (his children) is the concatenation of 'wlidat' (children) + 'ou' (suffix used to mark possession equivalent to his). Moreover, in MA dialect the agglutination is used also to combine particles with the main word, like in 'fl7ayat' (in the life) the letter 'f' (in) is a preposition concatenated with 'l' (the) a definite article and '7ayat' (life) a noun.

These linguistic features of written dialect are the main source of variations and non-uniformity of MA text in SM. In the next sections, we will present our solution to normalize dialect word forms and hence increase this text uniformity.

## 4 Data extraction

The used data was gathered from YouTube video comments extracted using YouTube API. YouTube is the most popular SM platform in Morocco, used by 44.48% of the population[7]. These videos have been selected using keywords related to various topics such as politics, sport, art, cooking, comedy, and others. For instance, for cooking we use key words like 'شهيوات مغربية', 'بسطيلة', 'حليمة الفيلالي'. The goal is to capture a wide range of words from different domains and thus ensure a large coverage of MA vocabulary. The collected corpus contains about 500K sentences.

---

[5] Both Classical Arabic and Standard Arabic.
[6] G*emination* or consonant lengthening is an articulation of a consonant for a longer period of time than that of a singleton consonant (from Wikipedia).
[7] https://gs.statcounter.com/social-media-stats/all/morocco

## 5   Data preprocessing

Before starting the normalization steps, we conduct a set of pre-processing to prepare our corpus for word embedding. The first one is the selection of Latin script comments because the raw corpus contains also Arabic script comments, then duplicated comments or containing only numbers or just one word are removed and any repetitions of more than three letters are reduced back to two letters. In addition, we removed all punctuations, hashtags, URLs, at mentions, emoticons, symbols, and full number strings. Finally, we have substituted the numbers within words by their equivalent letters to meet those used in the dialect dictionary (see Table 1). Except for '7' (equivalent of 'ح') and '3' (equivalent of 'ع') because we do not have their correspondent letters in Latin script, and we lowercased the overall text. After these pre-processing, the resulted corpus consists of 160 651 sentences and 242 277 unique words.

| Numbers/Letters | Equivalent letters |
|---|---|
| 2 | a |
| 6 | t |
| 4 / 8 | gh |
| 5 / x | kh |
| 9 | q |

Table 1. Conversion rules from numbers to letters as observed in our YouTube corpus

## 6   Dialect Normalization

Our system for dialect normalization is based on two steps. Namely, transliterations extraction and transliterations selection. Starting from three word embedding models and a MA dialect dictionary that will serve as the lexicon of canonical word form. For each canonical word in this dialect dictionary, we first extract the most semantically similar words from the vocabulary produced by the word-embedding models. Semantic similarity aims to select the nearest neighbors to each canonical word based on their context. Then, from these selected words, we select the most lexically similar ones to the canonical word. The lexical similarity aims to select the most similar transliterations to the canonical word in terms of surface form. We will describe these processing steps in detail in the following sections.

### 6.1   Dialect dictionary

The canonical form considered for our normalization task was first based on a collection of MA dialect dictionaries of nouns, verbs and adjectives (Jaafar, 2012). This dictionary consists of 14548 entries that we have converted from an adapted IPA (International Phonetic Alphabet) transcription to match the Latin script employed by Social Media users. the different adopted conventions are listed in Table 2. Then, we have extended this dictionary to include some special words to social media (e.g. 'tagini' (tag me)). However, by checking the collected dictionary, we found that, only 16% of dictionary words are present in the word embedding models vocabulary. In fact, today, the use of most of these dictionary words are not common in youth-run SM. To overcome this problem, we have semi-automatically collected a set of words from the models vocabulary while focusing on useful words that can capture other transliterations. We mention that we consider borrowed word from other languages as neology and we include them as new entries in MA dictionary. For instance, the word 'stationi' (to park) is borrowed from French 'stationner'. The final dictionary is of size 2502 canonical words.

| Adapted API script | Latin script | MA phoneme | API symbol |
|---|---|---|---|
| ḥ | 7 | ح | ħ |
| ḑ/ đ | d | ض | dˤ |
| ε / Ɛ | 3 | ع | ʕ |
| ġ | gh | غ | ɣ |
| ħ | h | ه | h |
| ḫ / x | kh | خ | x |
| ḻ | l | ل (geminated) | l |
| r̠ / r̥ / ř | r | ر | r |
| ṣ | s | ص | sˤ |
| š / ṣ̌ | ch | ش | š |
| ṭ / ṱ | t | ط | tˤ |
| ž | j | ج | ʒ |
| ẓ / ż | z | ز | z |
| â | a | ا | ʔ / a |
| ə | e | - | - |
| î | i | ي | i |
| û | ou | و | u |

Table 2. Conversion rules from adapted API to Latin script used for MA dialect in SM and its equivalent in MA phoneme with the associated API symbol

### 6.2 Word Embedding model generation

We use three word-embedding models, namely, word2vec CBOW (continuous bag of words) and Skip-gram (Mikolov et al., 2013) and the third one is FastText (Bojanowski et al., 2017). According to Mikolov et al. both architectures CBOW and Skip-gram work well in semantic tasks. In addition, Skip-gram is efficient in presenting infrequent words, unlike CBOW that has better representations for more frequent words. FastText is also able to better model infrequent words. Therefore, these models can complement each other and if we combine their outputs, we can expand the coverage of the detected transliterations for each canonical word form. For the configuration of these models, we use two as minimum count for each word occurrences in the corpus to capture rare words forms with a context window of seven words from the left and right sides.

### 6.3 Normalization and transliteration word forms Association

The normalization starts with the extraction of the nearest neighbors of each word in the MA dictionary (normalization word form) based on the semantic similarity score using the cosine distance between the dimensional vector of each word in the model and each canonical word. We have used the class most_similar of Gensim framework and we have fixed the list size parameter to twenty. Because we found that lower values capture few words and higher values capture a lot of noise. The same finding has been noted by Htait and Bellot (2018).

The second stage in this process is lexical similarity where we extract the nearest words to the canonical one (from MA dictionary) according to its surface form. We do so by measuring the similarity between each canonical word and the set of extracted words in the previous stage. Then, we select the words that have a similarity score higher than a threshold value. This value has been defined empirically and we have fixed it at 70%. After several experiments (more details will be given in Sec. 7), we have observed that the smaller is the threshold, the larger is the coverage but many undesirable words are selected. However, higher threshold values eliminate a considerable part of word transliterations certainly in the case of word agglutination. For example, we cannot capture and normalize 'lmagrib' (Morocco) to 'maghrib' if the threshold is up to 70%.

To compute the lexical similarity score, we have used different measures. The first one is based on sequence matching with vowels and double consonants removing. We removed vowels from words because as we have mentioned, most of writing variations are related to the use of different vowels. For example, the word 'ya3tek' (give you) can be written in 13 different manners: ya3tek, ye3tek, yaatik, ya3tik, yatek, yaatek, yaetik, y3atik, ya3tike, ytik, yi3tik, ya3atik, ye3tik. Therefore, by suppressing vowels and keeping consonants we can catch a large part of word transliterations. Moreover, we reduced back all consecutive double consonants that are used to mention gemination because it is not respected by all SM users. For instance, the word 'allah' (god) can be written 'alah' by some users.

The second measure used for lexical similarity is sequence matching with Soundex adapted to MA dialect phonetic rules as shown in Table 3.

Finally, we have used the other lexical similarity measures employed in related works. Our goal is to show the effectiveness of each measure in capturing different word' transliterations. In Table 4, we list the different formula used for scoring functions in each study. The performance of these measures will be given in the evaluation section.

| MA Soundex phonetic rules | Conversion |
|---|---|
| b, f, m, p, v, w | 1 |
| d, t, l, n | 2 |
| s, z | 3 |
| j, y, ch | 4 |
| r, kh, gh | 5 |

Table 3. MA phonetic rules used for Soundex

| Approaches | Semantic similarity | Lexical similarity |
|---|---|---|
| Sridhar (Sridhar, 2015) | Cosine similarity $$= \frac{\sum_{i=0}^{D} u_i \times v_i}{\sqrt{\sum_{i=0}^{D}(u_i)^2 \times \sum_{i=0}^{D}(v_i)^2}}$$ | $$\text{lexical similarity}(s_1, s_2) = \frac{\text{LCSR}(s_1, s_2)}{ED(s_1, s_2)}$$ $$\text{LCSR}(s_1, s_2) = \frac{\text{LCS}(s_1, s_2)}{Max\ Lenght(s_1, s_2)}$$ LCSR = Longest Common Subsequence Ratio (Melamed, 1995) <br> LCS = Longest Common Subsequence <br> ED8 = Edit distance between the two strings. |
| Bertaglia (Bertaglia and Nunes, 2016) | Cosine similarity (same formula) | $$\text{lexical similarity}(s_1, s_2) = \begin{cases} \frac{\text{LCSR}(s_1, s_2)}{MED(s_1, s_2)}, & if\ MED(s_1, s_2) > 0 \\ \text{LCSR}(s_1, s_2), & \text{otherwise} \end{cases}$$ $$\text{LCSR}(s_1, s_2) = \frac{\text{LCS}(s_1, s_2) + DS(s_1, s_2)}{Max\ Lenght(s_1, s_2)}$$ MED($s_1,s_2$) = ED($s_1,s_2$) - DS($s_1,s_2$) <br> MED = Modified edit distance <br> DS = Diacritical symmetry between s1 and s2 <br> N.B: As diacritics does not exist in our case, so DS = 0 and the lexical similarity formula will be the same as Sridhar one. |
| Htait [9] (Htait et al., 2018) | Cosine similarity (same formula) | Sequence Matching with a score of 50% |

Table 4. Scoring functions for both semantic and lexical similarities used in normalization approaches based on word embedding

---

[8] For English, the edit distance computation was modified to find the distance between the consonant skeleton of the two strings $s_1$ and $s_2$.
[9] Our implementation was initially based on the open code provided by Htait: https://github.com/OpenEdition/NormAFE

We have executed the same normalization process for the three generated word-embedding models. We have observed that each model can capture a set of different transliterations. Therefore, we have decided to merge the produced lexicon in each case. The resulting combinations (transliteration, normalization form) are then grouped together to form a single MANorm normalization dictionary.

## 7 Evaluation

The goal of the evaluation of MANorm dictionary is to test the quality of the normalization lexicon. Since this is the first work on MA dialect normalization, we do not have a gold standard or a reference to evaluate the performance of the produced lexicon. Therefore, it is impossible to proceed to an automatic evaluation, thus we have created our own reference. We first combined the outputs dictionaries of the three models, which produced a normalization dictionary of 3057 entries (transliteration, normalization form). We have then, validated each entry based on a human judgment by checking manually the correct association between each word transliteration and its canonical form. This operation resulted in a normalization reference dictionary of 2225 correct entries. Examples of result are given in Table 5.

| Canonical word form (MA dialect dictionary) | Transliterations (corpus) |
|---|---|
| awel (the first) | awl, awwal, awle, aaal, awale |
| choukran (thank you) | chokran, chokrane, chkran, khokran, chokrn |

Table 5. Normalization result examples

We have used precision and coverage as evaluation metrics rather than the common information retrieval ones (precision, recall and F-score). Because we are not able to measure the recall, which is the ratio between the number of the provided relevant results, and the total of relevant results that we have to provide. Since we do not know exactly how many transliterations, we can normalize by each given canonical word form, we cannot define the total value of relevant results. Therefore, instead of recall, we measure the coverage of the MA dictionary words. The coverage represents the ratio of useful canonical words or the words that were able to catch some other transliterations over the total of MA dictionary words.

To gain more insight into the performance differences of the three word embedding models individually, we calculate their coverages, and we measure their precisions by comparing each model with the created reference normalization dictionary. The results are shown in Table 6.

By looking at Table 6, we first observe that CBOW model achieved the best precision followed by Skip-gram, but their coverage is still very low. Second, FastText performance was the worst. Finally, it can be seen that after combining the three models' outputs we achieved the best coverage with a huge margin while keeping a good precision. From these results, we can conclude that models' combination allows to counter the imbalance between the high precision and low coverage of separated models.

We also evaluate the different scoring functions of lexical similarity used in other related works (with a threshold of 70%) and we report the results in Table 7. These results are related to the three merged models.

Table 7 shows that Lexim (used in Sridhar and Bertaglia) outperforms all the other scoring functions in term of precision but its coverage was the worst. In term of coverage, sequence matching with Soundex achieved the best coverage among all the other scoring functions. However, sequence matching with vowels and double consonants removing have balanced the precision and coverage scores. For this reason, we consider this last lexical similarity scoring function in MANorm dictionary generation.

To prove our choice of the threshold value for lexical similarity scoring we conduct a series of experiments. The obtained results are reported in Table 8. The main conclusion we can draw from this table is that higher threshold values give better precision but lower coverage. By raising the threshold from 60% to 80%, the precision improves considerably but the coverage decreases drastically. Thereby, in MANorm, we used the medium threshold 70% that balances between precision and dictionary coverage.

| Models | Precision | Coverage |
|---|---|---|
| Wrod2vec CBOW | 0.815 | 0.300 |
| Word2vec Skip-Gram | 0.775 | 0.280 |
| FastText | 0.414 | 0.163 |
| Merged dictionary (CBOW + Skip-gram + FastText) | 0.704 | 0.463 |

Table 6. Word embedding models performance

| Lexical similarity scoring function | Precision | Coverage |
|---|---|---|
| Lexim (used in Sridhar and Bertaglia) | **0.785** | 0.300 |
| Sequence matching (used in Htait) | 0.671 | 0.427 |
| Sequence matching + Soundex | 0.486 | **0.578** |
| Sequence matching + vowels and double consonants removing | 0.704 | 0.463 |

Table 7. Lexical similarity scoring functions comparison

| Threshold value | Precision | Coverage |
|---|---|---|
| 60% | 0.414 | 0.706 |
| 65% | 0.428 | 0.697 |
| **70%** | **0.704** | **0.463** |
| 75% | 0.703 | 0.462 |
| 80% | 0.744 | 0.424 |

Table 8. Threshold value influence on merged models precision and coverage

While validating MANorm, we have observed that agglutination which is mainly related to word inflection, is a real source of ambiguity because it opens the door for a high number of possible transliterations. As our main concern is to reduce spelling variation, in case of inflected word forms (e.g. conjugated verbs, nouns and adjectives plural and feminine forms), we consider word lemma or the nearest inflection form as the correct normalization. In MA dialect, the lemma of verbs is the past tense of the third person singular and for nouns and adjectives, the lemma is the masculine singular form. For instance, as shown in Table 9, the normalized form of the adjectives 'saknin' and 'sakna' is the lemma form 'saken'. Other examples are presented in Table 9.

Regarding normalization errors, they are related to different reasons. In some cases, we have observed, that one transliteration (input) can be assigned to several canonical forms. In such a case, throughout validation, we consider correct the canonical form that is closest in meaning according to the transliteration context that we find in the corpus. For example, the transliteration '7amad' has been assigned to '7amd' (praise) and '7amed' (sour) but according to their context the correct form is the second one.

We have encountered another issue related to agglutination, for example, the transliteration 'wqaaf' was affected to two canonical forms 'wqef' (hold on) and 'awqaf' (Islamic endowments). However, the word 'awqaf' used in the corpus is a concatenation of 'a' and 'wqaf' that means hold on (in a strong way). By checking their contexts in the corpus, we found that 'wqaaf' has the same meaning as 'wqef' that we therefore consider as correct.

Other errors are due to the fact that the phonemes transliteration conventions adopted in our dictionary do not always meet those used in the corpus. For instance, the noun '7aj' (pilgrimage) was assigned to the verb 'haj' (to rave) where 'h' was used as [ħ] by some users. However, in MA dictionary we use '7' for this phoneme and 'h' to represent [ħ]. This problem is partly due to that during lexical similarity we do not differentiate between '7' and 'h', because doing so we can gather a large number of different transliterations.

There are cases where a transliteration can be given an inappropriate canonical form because both are used in the same context, and are lexically close to each other (similarity > 70%), although, they belong to different words. For example, '3id' (Eid/religious celebration) was assigned to 'sa3id' (happy) as canonical form.

| Word POS tag | Inflected form | Lemma/Nearest inflection |
|---|---|---|
| Adjective | saknin (they are living) <br> sakna (she is living) | Lem. masculine singular: <br> saken (he is living) |
| Noun | klamo (his words) <br> klamek (your words) | Lem. masculine singular: <br> klam (words) |
| Verb | 3aqo (they realized) <br> 3aqna (we realized) | Lem. past tense, third person singular: <br> 3aq (he realized) |
| Verb | bakatni (she made me cry) <br> bakitini (you (singular) made me cry) <br> bakitona (you (plural) made me cry) | Lem. past tense, first person singular: <br> bkite (I cried) |
| Verb | 3ajbatni (I liked it) <br> 3jbatni (I liked it) <br> 3jebni (I liked it) <br> kat3jabni (I like) <br> kay3jabni (I like) | Nearest inflection: <br> 3jabni (I liked it) |

Table 9. Inflected word normalization

Finally, even if errors are quite common in MaNorm, it is still an interesting attempt to normalize dialect spelling variation. We are confident that with larger corpus and dictionary we can further more improve its performance.

## 8 Conclusion

Written dialect is a phonetic transliteration of spoken words that does not follow any standard orthography. It is mostly used in SM where each user improvises his own spelling. As a result, for each single word we find a mixture of spellings. Before performing any NLP task, it is mandatory to transform these dialect words written differently to one normalized form. In this work, we present our solution for MA dialect normalization based on semantic similarity using word-embedding and lexical similarity. We use three word-embedding models that we combine their outputs to form one normalization dictionary MANorm. In the resulting dictionary, we find the matching between each word from the corpus and the most similar word form from the MA dictionary. The merge allows us to make a compromise between precision and coverage. The evaluation of the normalization dictionary shows the good performance of this solution. However, other improvements can be done with a larger corpus, especially to increase MA dictionary coverage. As next direction, we will employ the same technique to normalize MA words written in Arabic script after building the appropriate dictionary for canonical forms.